\documentclass[runningheads]{llncs}

\usepackage[FINAL,year=2024,ID=10982]{eccv}

\usepackage{eccvabbrv}

\usepackage{graphicx}
\usepackage{booktabs}

\usepackage[accsupp]{axessibility}  

\usepackage{times}
\usepackage{epsfig}
\usepackage{graphicx}
\usepackage{amsmath}
\usepackage{amssymb}
\usepackage{multirow}
\usepackage{diagbox}
\usepackage{xcolor}
\usepackage{booktabs}
\usepackage{svg}
\usepackage[lined,boxed,commentsnumbered,ruled]{algorithm2e}

\usepackage[pagebackref,breaklinks,colorlinks,citecolor=eccvblue]{hyperref}

\usepackage{orcidlink}

\begin{document}

\title{Anisotropic Neural Representation Learning for High-Quality Neural Rendering} 


\author{$\text{Yifan Wang}^*$ \and
$\text{Jun Xu}^*$ \and
Yi Gong \and Yuan Zeng  }

%

\authorrunning{F.~Author et al.}

\institute{Southern University of Science and Technology  \\
}

\maketitle
\footnotetext{$^*$ Authors contributed equally to this work.}

\begin{abstract}
Neural radiance fields (NeRFs) have achieved impressive view synthesis results by learning an implicit volumetric representation from multi-view images. To project the implicit representation into an image, NeRF employs volume rendering that approximates the continuous integrals of rays as an accumulation of the colors and densities of the sampled points. Although this approximation enables efficient rendering, it ignores the direction information in point intervals, resulting in ambiguous features and limited reconstruction quality. In this paper, we propose an anisotropic neural representation learning method that utilizes learnable view-dependent features to improve scene representation and reconstruction. We model the volumetric function as spherical harmonic (SH)-guided anisotropic features, parameterized by multilayer perceptrons, facilitating ambiguity elimination while preserving the rendering efficiency. To achieve robust scene reconstruction without anisotropy over-fitting, we regularize the energy of the anisotropic features during training. Our method is flexible and can be plugged into NeRF-based frameworks. Extensive experiments show that the proposed representation can boost the rendering quality of various NeRFs and achieve state-of-the-art rendering performance on both synthetic and real-world scenes. 
  \keywords{Neural radiance fields \and Anisotropic implicit representation \and Neural rendering}
\end{abstract}

\begin{figure}[h]
	\begin{center}
		\includegraphics[width=1\linewidth]{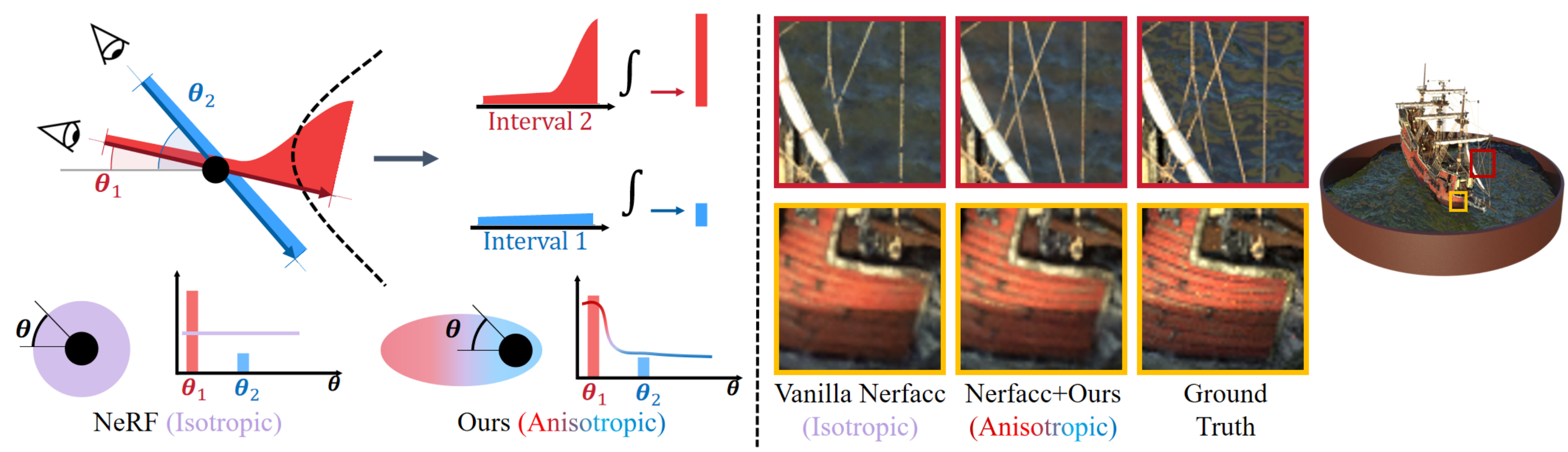}
	\end{center}
	\caption{Left: Vanilla NeRF uses point sampling and view-independent functions to estimate $\mathbf{\sigma}$ and $\mathbf{e}$, resulting in directional ambiguity when representing intervals. To eliminate the ambiguity, we introduce anisotropic functions to model the integration along various directions. Right: Our anisotropic neural representation enables to assist Nerfacc\ \cite{li2023nerfacc} in capturing more geometric details and realistic textures.}
	\label{fig:first}
\end{figure}
\section{Introduction}
\label{sec:intro}

Neural radiance field (NeRF)\ \cite{mildenhall2021nerf} models the 3D scene geometry and view-dependent appearance by two cascaded multi-layer perceptrons (MLPs) and uses volume rendering to reconstruct photorealistic novel views. The advent of NeRF has sparked a flurry of work on neural rendering and has opened the way to many applications\ \cite{chen2022mobilenerf, chen2021mvsnerf, xu2022point, xing2022mvsplenoctree, barron2021mipnerf, yang2023freenerf, wang2023hyb}. One of the key underpinnings of NeRF is differentiable volume rendering which facilitates learning the MLP-based radiance field solely from a 2D photometric loss on synthesized images in an end-to-end manner. However, NeRF still suffers from blurring and aliasing when reconstructing complex scenes. Specifically, since NeRF is done by sampling a set of points along a ray and approximating the piece-wise continuous integration as an accumulation of the estimated volumetric features of sampled intervals, MLP could only be queried at a fixed discrete set of positions and the same ambiguous point-sampled feature is used to represent the opacity at all points with the interval between two samples, leading to ambiguity in neural rendering. The features of samples can be very different across different rays cast from different viewing directions. Therefore, simultaneously supervising these rays to produce the isotropic volumetric features can result in artifacts like fuzzy surfaces and blurry texture, as shown in Fig. \ref{fig:first}.

To overcome the limitations of the neural scene representation and enhance rendering quality of NeRF, several works\ \cite{barron2021mipnerf,barron2022mipnerf360, isaac2023exact, hu2023tri, barron2023zip} introduce shape-based sampling into the scene representation. By embedding novel spatial parameterization schemes, such as Gaussian ellipsoid, frustum, and sphere, into the encoding, these models can reduce representation ambiguity and improve rendering quality with less blurring and aliasing artifacts. However, the representation ambiguities still exist in their radiance fields, since the directional intervals used for rendering are parameterized by the non-directional features. A straightforward solution is to take the viewing direction as input of the first MLP, enabling to represent the scene geometry with view-dependent features. While this method can reduce directional ambiguity in representation, the radiance field is now a high-degree view-dependent function without regularization, and it is prone to over-fitting to the appearance of training images but fails to capture the correct geometry\ \cite{zhang2020nerf++}.  

In this paper, we propose a novel radiance field representation to diverse a different model for the density and appearance in neural rendering, leading to better reconstruction of the scene's geometry and appearance while producing photorealistic novel views. Instead of introducing a different spatial sampling strategy and parameterizing sampled-shapes, we model the density and latent features at a location as view-dependent functions using spherical harmonic (SH) basis. The spherical harmonics can be evaluated at arbitrary query viewing directions to capture anisotropy. Although this can be done by converting an existing NeRF into such anisotropic representations via projection onto the SH basis functions, we simply modify the first MLP in NeRF to predict the geometry explicitly in terms of spherical harmonics, resulting in a compact and generalizable anisotropic neural representation. Specifically, we train the first MLP that produces coefficients for the SH functions instead of the density and latent features. The predicted values can later be directly used for appearance estimation and pixel rendering. 

Additionally, training the anisotropic neural representation using only a color reconstruction loss will cause strong anisotropy and suffer from shape-radiance ambiguity in rendering\cite{zhang2020nerf++}. Although existing regularization methods, such as Patch-based Consistency \cite{chen2023structnerf},warp-based loss \cite{zhang2022multi}, depth or multi-view multi-view stereo prior \cite{wang2022neuris, wu2023s, wang2022neuralroom, bian2022nopenerf} can be a solution, they rely on the geometric prior and structural output and significantly increase the computational cost. We aim to introduce an effective anisotropy regularization according to our SH function-based representation. To this end, we decompose the anisotropic elements from the predicted density and latent features and employ point-wise operation to penalize the anisotropy in the geometry representation. This improves the memory and computational efficiency of our method and allows us to render high-quality novel views. Moreover, our anisotropic neural representation can be used as a sub-model to replace the first MLP in various NeRFs, resulting in anisotropic geometric features and better rendering quality.   

We extensively evaluate the effectiveness and generalizability of our anisotropic neural representation on benchmark datasets including both synthetic and real-world scenes. Both quantitative and qualitative comparison results demonstrate that plugging our anisotropic neural representation can further improve the rendering quality of various NeRFs. Our contributions are summarized as follows:
\begin{itemize} \itemsep0.75pt
	\item A novel anisotropic neural representation is proposed to model the scene geometry by view-dependent density and latent features. It effectively reduces directional ambiguity in neural rendering and results in better geometric and appearance reconstruction and rendering quality. 
	\item A modified NeRF network is trained to predict view-dependent geometry in terms of spherical basis functions,  which is flexible and generalizable to various existing NeRFs.
	\item A point-wise anisotropy regularization loss enables highly efficient view-dependent penalties during model training to avoid over-fitting.
\end{itemize}

\section{Related Work}
\label{sec:related}
\paragraph{Scene representations for Novel view synthesis.}
Synthesizing views from a novel viewpoint is a long-standing problem in computer vision and computer graphics. Traditional methods typically synthesize novel views from a set of images\ \cite{buehler2023unstructured, chaurasia2013depth, chaurasia2011silhouette, debevec2023modeling, sinha2009piecewise} or light fields\ \cite{levoy2023light, srinivasan2017learning, wood2023surface}. Although these methods work well on dense input images, they are limited by the sparse inputs and the quality of 3D reconstruction. To synthesize novel views from a sparse set of images, some methods leverage the geometry structure of the scene\ \cite{penner2017soft, goesele2010ambient}. Recent advances in deep learning have facilitated the use of neural networks for estimating scene geometry, such as voxel grids\ \cite{sitzmann2019deepvoxels, he2020deepvoxels++, lombardi2019neural, rematas2020neural}, point clouds\ \cite{song2020deep, le2020novel, achlioptas2018learning,wang2019mvpnet}, and multi-plane images\ \cite{choi2019extreme, zhou2018stereo}. Although these discrete representation methods can improve the rendering quality of novel views, the estimation of the scene geometry is not accurate enough for high-resolution scenes. 

By modeling the scene geometry and appearance as a continuous volume, neural radiance fields (NeRFs) \cite{mildenhall2021nerf} have achieved state-of-the-art novel synthesis effects. Specifically, NeRF maps the input coordinate to density and radiance scene values and uses volume rendering\ \cite{max1995optical} to synthesize the images. In addition, various schemes have been introduced to improve the robustness of NeRF to few-shot inputs\ \cite{yang2023freenerf, wang2023sparsenerf, niemeyer2022regnerf, kim2022infonerf}, anti-aliasing\ \cite{barron2021mipnerf, barron2022mipnerf360,  barron2023zip, hu2023tri}, handle dynamics\ \cite{martin2021nerf, chen2022hallucinated, mildenhall2022nerf, xin2021hdr} and speed up rendering\ \cite{Chen2022ECCV, fridovich2022plenoxels, muller2022instant, sun2022direct, han2024volume}, etc. Our method is more closely related to anti-aliasing NeRFs, which adopt interval-dependent features to assist the MLP in capturing more accurate geometry and reducing blurring artifacts in novel view synthesis. In contrast, we introduce a plug-and-play anisotropic neural representation that enables it to be plugged into various NeRFs to alleviate ambiguity and improve rendering quality.   
\par   
\paragraph{Anti-ambiguity in neural rendering.}
Volume rendering is an important technique with a long history of research in the graphics community. Traditional graphics rendering methods include studying ray sampling efficiency and data structures for coarse-to-fine hierarchical sampling. Recent NeRF and its succeeding works have shown impressive results by replacing or augmenting the traditional graphics rendering with neural networks. The volume rendering used in these works is approximated to a discrete accumulation under the assumption of a piece-wise constant opacity and color, enabling to learn the NeRF-based implicit representations. However, the piece-wise constant assumption results in rendering results that are sensitive to the sampled points as well as the cumulative density function of the distribution of the sampled interval, introducing ambiguous features and aliasing artifacts in NeRF renderings. 

To address these challenges, recent works have explored super-sampling or pre-filtering techniques. Super-sampling is done by casting multiple rays per pixel to approach the Nyquist frequency. Although this strategy works well for eliminating ambiguity, it is computationally expensive. Pre-filtering techniques are more computationally efficient, since the filtered versions of scene content can be pre-computed ahead of time. Recently, pre-filtering has been introduced into neural representation and rendering to reduce ambiguity and aliasing artifacts\ \cite{blackman1958measurement}. Mip-NeRF\ \cite{barron2021mipnerf} samples the cone instead of rays to consider the shape and size of the volume viewed by each ray and optimizes a pre-filtered representation of the scene during training. With sampled shape-dependent inputs, the MLP of Mip-NeRF can capture various volumetric features to mitigate ambiguity, resulting in a high-quality multi-scale representation and anti-aliasing. Mip-NeRF 360\ \cite{barron2022mipnerf360} extends Mip-NeRF with a novel distortion-based regularizer to tackle unbounded scenes. Zip-NeRF\ \cite{barron2023zip} adopts multi-sampling to approximate a cone with hash encoding. Tri-Mip\ \cite{hu2023tri} leverages multi-level 2D mipmaps to model the pre-filtered 3D feature space and projects parameterized spheres on three mipmaps to achieve anti-aliasing encoding. Although the shape and size of volumes at different scales can be fitted by introducing various sampling techniques in NeRF, shape-based features still exist in ambiguity since the non-directional shape intervals are used in accumulation for pixel rendering. Our work draws inspiration from the early work on anisotropic volume rendering\ \cite{schussman2004anisotropic} and is the first to model scene geometry as an anisotropic neural representation based on spherical harmonics for volume rendering.


\section{Preliminaries}
\label{sec:prelim}

\paragraph{Neural Radiance Fields.} Given a 3D position $\mathbf{x}$ and a 2D viewing direction $\mathbf{d}$, NeRF\ \cite{mildenhall2021nerf} first uses a multilayer perceptron (MLP) parameterized by weights $\theta$ to predict the density $\mathbf{\sigma}$ and an intermediate vector $\mathbf{e}$ from the input position $\mathbf{x}$: $(\mathbf{\sigma}, \mathbf{e})=\mathcal{F}_{\theta}(\mathbf{x})$. Then, a second MLP parameterized by weights $\phi$ is employed to estimate the color $\mathbf{c}$ from the direction $\mathbf{d}$ and the vector $\mathbf{e}$: $\mathbf{c}=\mathcal{F}_{\phi}(\mathbf{d}, \mathbf{e})$.\par  

\paragraph{Volume Rendering.} 
The color of a pixel in NeRFs can be rendered by casting a ray $\mathbf{r}(t)=\mathbf{o}+t\mathbf{d}$ from the camera origin $\mathbf{o}$ through the pixel along the direction $\mathbf{d}$, where $t$ is the distance to the origin. The pixel's color value can be computed by integrating colors and densities along a ray based on the volume rendering\ \cite{max1995optical}:
\begin{equation}
	\hat{C}(\mathbf{r})=\int_{0}^{\infty} T(t)\sigma(\mathbf{r}(t))\mathbf{c}(t)\, dt,
	\label{ren}
\end{equation}
where $T(t)=\exp\left(-\int_{0}^{t}\sigma(\mathbf{r}(s))\, ds\right)$ represents occlusions by integrating the differential density between $0$ to $t$. 

Since the volume density $\mathbf{\sigma}$ and radiance $\mathbf{c}$ are the outputs of MLPs, NeRF rendering methods approximate this continuous integral using a sampling-based Riemann sum instead\ \cite{riemann1854hypotheses}. Within the near and far bounds, $t_{n}$ and $t_{f}$ of the cast ray, a subset of points on the ray is sampled in a near-to-far order. Let $\mathbf{t}_{N}={t_{1}, \cdots, t_{N}}$ be $N$ samples on the ray that define the intervals, e.g., $I_{i}=[t_{i}, t_{i+1}]$ is the $i$th interval, and $I_{0}=[0, t_{1}]$, $I_{N}=[t_{N}, \infty]$. The volume density for particles along the interval $I_{i}$ is predicted under the assumption that opacity is constant along each interval, which indicates $\sigma(\mathbf{r}(t))=\sigma(\mathbf{r}(t_{i})), \forall{t}\in[t_{i}, t_{i+1}]$ for particles of constant radius and material. We denote $\sigma(\mathbf{r}(t_{i}))=\sigma_{i}$ for notation convenience. Under this assumption, the rendered color can be written as an approximation of the $N$ sampled points:


\begin{equation}
	\begin{aligned}
		\begin{split}
			\hat{C}(\mathbf{r}) =\sum_{i=0}^{N}\mathbf{c}_{i}T_i(1-\exp(-\sigma_{i}(t_{i+1}-t_{i}))),
		\end{split}
		\label{ren}
	\end{aligned}
\end{equation}

where $T_{i}=\exp\left(-\sum_{j=0}^{i-1}\sigma_{j}(t_{j+1}-t_{j})\right)$ represents the transmittance accumulated along the ray until the $i$th sample. For more detailed derivation, please refer to\ \cite{max2005local}. The NeRF model is optimized by minimizing the $L_{2}$ reconstruction loss between the ground truth and synthesized images, which can be expressed as follows:
\begin{equation}
	\mathcal L_{recon} = \frac{1}{|\mathcal{R}|}\sum_{\mathbf{r}\in \mathcal{R}} \left\|\hat{C}  (\mathbf{r}) - C(\mathbf{r})\right\|_{2}^{2},
\end{equation}


where $\mathcal{R}$ is a set of rays sampled during training, and $C$ is the ground truth pixel color value. \par

\paragraph{Limitations.} 

While images can be efficiently rendered using NeRF renderings, it is non-trivial to represent directional intervals with the point-sampled features. In addition, leveraging the same predicted features to represent the distribution along the interval between two samples can hardly capture the correct geometry of different rays cast from different directions. For example, under piece-wise constant color, the contribution $\hat{C}_{i}(\mathbf{r})$ of the $i$th interval to $\hat{C}(\mathbf{r})$, i.e., the volume rendering integral of the interval $I_{i}$, can be written as\ \cite{max1995optical,mildenhall2021nerf}:
\begin{equation}
	\begin{aligned}
		\begin{split}
            \hat{C}_{i}(\mathbf{r})&=\int_{t_{i}}^{t_{i+1}}\mathbf{c}(t)\sigma(\mathbf{r}(t))e^{-\int_{0}^{t}\sigma(\mathbf{r}(s))ds} dt       \\
            &=\mathbf{c}_{i}e^{-\int_{0}^{t_{i}}\sigma(\mathbf{r}(s))ds}(1-e^{-\int_{t_{i}}^{t_{i+1}}\sigma(\mathbf{r}(s))ds})            \\
            &=\mathbf{c}_{i}T(t_{i})(1-e^{-\int_{t_{i}}^{t_{i+1}}\sigma(\mathbf{r}(s)ds}).
		\end{split}
	\end{aligned}
 \label{ren2}
\end{equation}
To achieve efficient rendering, NeRF uses point-sampled density $\sigma_{i}$ to approximate the distribution along the directional interval ($\int_{t_{i}}^{t_{i+1}}\sigma(\mathbf{o}+s\mathbf{d})ds/(t_{i+1}-t_i)$) and simplifies the rendering computation in equation (\ref{ren2}) as $\mathbf{c}_{i}T_i(1-\exp(-\sigma_{i}(t_{i+1}-t_{i})))$ in equation (\ref{ren}). This indicates that the integral of $\sigma(\mathbf{o}+s\mathbf{d})$ is related to the viewing direction, which has also been studied in\ \cite{schussman2004anisotropic} for volume rendering. Approximating the continuous integral of the interval as an accumulation of the isotropic features of sampled points has no guarantee for correct approximation. Fig.\ \ref{fig:first} shows that the vanilla NeRF using an isotropic representation has limited rendering quality and suffers from blurring artifacts. This insight motivates our anisotropic neural representation.  

\section{Method}
\label{sec:method}
\begin{figure}[t]
	\begin{center}
		\includegraphics[width=1\linewidth]{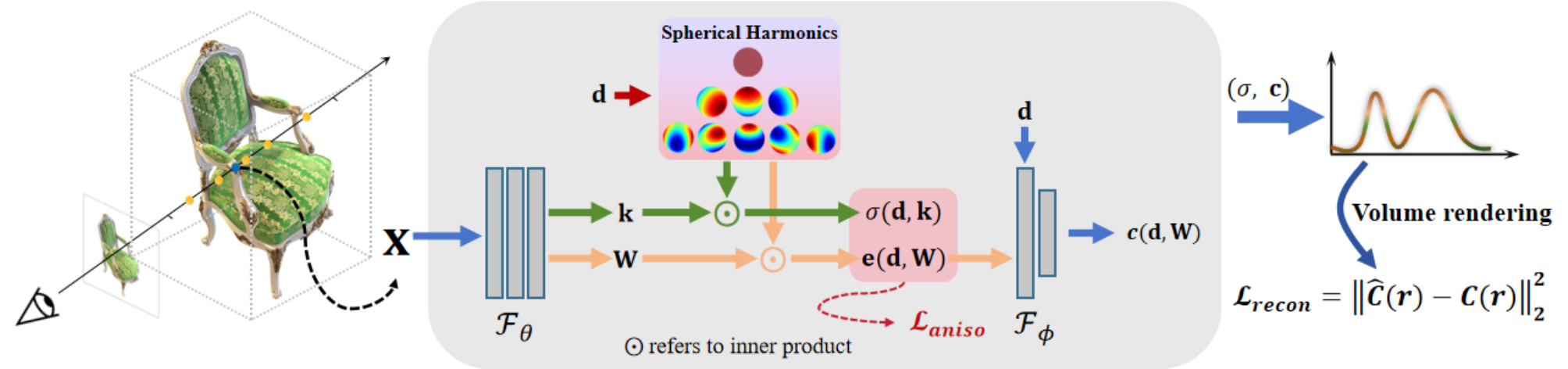}
	\end{center}
	\caption{Overview of our anisotropic neural representation. In contrast to output isotropic $\sigma$ and $\mathbf{e}$ directly as in vanilla NeRF, we composite anisotropic features by learning the SH coefficients predicted from MLP $\mathcal{F}_{\theta}$. During training, the model is optimized end-to-end by minimizing a joint loss of $\mathcal{L}_{recon}$ and $\mathcal{L}_{aniso}$.}
	\label{fig:architure}
\end{figure}
Given a set of scenes with a collection of images and their camera parameters, we aim to learn an anisotropic neural representation for high-quality neural rendering.  In this section, we first represent the scene geometry and appearance using spherical harmonic-based implicit representations to capture the anisotropy of surfaces (Sec. \ref{subsec: An-re}). Then we introduce an anisotropy regularization to encourage a sparse anisotropic neural representation and summarize our overall training procedure (Sec. \ref{subsec: training}). 
\par 
\subsection{Anisotropic Neural Representation}
\label{subsec: An-re}
Fig. \ref{fig:architure} illustrates an overview of our method. While vanilla NeRF takes a 3D position $\mathbf{x}$ as an input of a density MLP $\mathcal{F}_{\theta}$ to estimate the scene geometry, including volume density $\mathbf{\sigma}$ and latent feature $\mathbf{e}$, we leverage both the 3D position and 2D viewing direction to introduce anisotropic features to represent the scene geometry. Although naively embedding the direction $\mathbf{d}$ into the density MLP $\mathcal{F}_{\theta}$ can realize anisotropic implicit geometry representation, this design will directly increase the dimension of input data and cause the model more easily fit the training data and estimate incorrect scene geometry\ \cite{zhang2020nerf++}. To capture anisotropy in scene representation, our method utilizes spherical harmonics (SH), which have been used to model Lambertian surfaces\ \cite{basri2003lambertian, ramamoorthi2001relationship}, or even glossy surfaces \cite{sloan2023precomputed}. We query the SH functions $Y_l^m$: $\mathbb{S}^{2}\mapsto \mathbb{R}$ at a viewing direction $\mathbf{d}$ and then fit the anisotropic neural representations by finding the corresponding coefficients. We use low-degree SH functions to compute ideal values of view-independent density and latent components, and high-degree SH functions for view-dependent components.   

For any sampled point $\mathbf{x}$ in the space, we adapt $\mathcal{F}_{\theta}$ to estimate the spherical harmonic coefficients $\mathbf{k}$ and $\mathbf{W}$, rather than the volume density and latent feature:

\begin{equation}
	(\mathbf{k}, \mathbf{W})=\mathcal{F}_{\theta}(\mathbf{x}),
\end{equation}
where the spherical harmonic coefficients $\mathbf{k}=\left(k_{l}^{m}\right)_{l: 0 \leq l \leq L}^{m:-l \leq m \leq l}$ is related to calculate the view-dependent volume density and $\mathbf{W}=[ \mathbf{w}_{1}, \cdots, \mathbf{w}_{K}]^{\top}$ consists of $K$ sets of SH coefficients that used to determine the $K$-dimensional latent feature vector. For $n\in \{1,\cdots, K\}$, we have $\mathbf{w}_n=\left(w_{nl}^{m}\right)_{l: 0 \leq l \leq L}^{m:-l \leq m \leq l}$. For $\mathbf{k}$ or $\mathbf{w}_n$, there are $(L+1)^{2}$ spherical harmonics of degree at the most $L$. The view-dependent density $\mathbf{\sigma}$ and latent feature $\mathbf{e}$ at position $\mathbf{x}$ are then determined by querying the SH functions $Y_l^m$ at the desired viewing direction $\mathbf{d}$:
\begin{equation}
	\sigma (\mathbf{d},\mathbf{k}) = \sum_{l=0}^{L} \sum_{m=-l}^{l} k_l^m Y_l^m(\mathbf{d}),  
	\label{eq:sh_1}
\end{equation}
and
\begin{equation}
	e_n (\mathbf{d},\mathbf{w}_n) = \sum_{l=0}^{L} \sum_{m=-l}^{l} w_{nl}^m Y_l^m(\mathbf{d}).
	\label{eq:sh_2}
\end{equation}
The equations (\ref{eq:sh_1}) and (\ref{eq:sh_2}) can be seen as the factorization of the density and latent feature with the isotropic and anisotropic SH basis functions, respectively. This eliminates the input of view direction to the density MLP and enables efficient generation of view-dependent geometry features. Then, a color MLP takes the inputs of the estimated latent feature $\mathbf{e}(\mathbf{d},\mathbf{W})$ and the direction $\mathbf{d}$ to predict the color value $\mathbf{c}$:
\begin{equation}
	\mathbf{c}(\mathbf{d}, \mathbf{W})=\mathcal{F}_{\phi}\left(\mathbf{d}, \mathbf{e}(\mathbf{d},\mathbf{W})\right).
\end{equation}

Given the estimated volume density $\mathbf{\sigma}(\mathbf{d}, \mathbf{k})$ and color value $\mathbf{c}(\mathbf{d}, \mathbf{W})$, a pixel color $\hat{C}(\mathbf{r})$ in the radiance field along the ray $\mathbf{r}$ can be predicted using the volume rendering in equation (\ref{ren}). Note that since the proposed scene geometry representation only adapts the original density MLP to learn SH coefficients and efficiently convert the input position into anisotropic features, making it easy to be plugged into various existing NeRF-based scene representation models and assists the models in capturing more precise geometries for high-quality novel view synthesis.
\par

\subsection{Training}
NeRF adopts pixel-wise RGB reconstruction loss $\mathcal{L}_{recon}$ to optimize view-independent geometry density and view-dependent color for scene reconstruction. However, it is difficult to optimize a correct geometry purely from the input RGB images, especially when view-dependent components are used in the geometry representation, since a high degree of anisotropy will be learned to fit the training images, leading to shape-radiance ambiguity and the degradation of rendering quality. To facilitate robust mapping under our anisotropic neural representation, we propose a point-wise anisotropy constraint, which penalizes the anisotropy of the model to mitigate shape-radiance ambiguity efficiently. 
\par

\label{subsec: training}
\paragraph{Anisotropy Regularization.} 
Without any regularization, the model is free to fit a set of training images by exploiting view-dependent anisotropic neural representation rather than recovering the correct geometry. The representation with strong anisotropy would generate shape-radiance ambiguity, resulting in blurring artifacts and incorrect geometries in rendering novel test views. We therefore introduce a new regularization method that suppresses the anisotropy in geometry representations.  

To apply regularization techniques for penalizing anisotropy, we first define the anisotropic features according to the SH basis functions. Since the $0$-degree SH function $Y_0^0(\mathbf{d})$ is view-independent, we remove the view-independent component from the estimated density $\mathbf{\sigma}(\mathbf{d},\mathbf{k})$ and latent feature $\mathbf{e}(\mathbf{d},\mathbf{W})$ to compute the view-dependent component: 
\begin{equation}
    \sigma ^{aniso} (\mathbf{d},\mathbf{k})=\sum_{l=1}^{L} \sum_{m=-l}^{l} k_l^m Y_l^m(\mathbf{d}),
\end{equation}
and 
\begin{equation}
	e_n^{aniso} (\mathbf{d},\mathbf{w}_n)=\sum_{l=1}^{L} \sum_{m=-l}^{l} w_{nl}^m Y_l^m(\mathbf{d}).
\end{equation}
We formulate our anisotropy regularization loss as:
\begin{equation}
	\label{eq:regu}
	\mathcal L_{aniso} = \frac{1}{N}\sum_{i=1}^{N}\left(\left\|\mathbf{\sigma}^{aniso}(\mathbf{d}_{i},\mathbf{k}_{i})\right\|^2_2+\left\|\mathbf{e}^{aniso}(\mathbf{d}_{i},\mathbf{W}_{i})\right\|^2_2\right),
\end{equation}
where $\mathbf{d}_i$, $\mathbf{k}_{i}$ and $\mathbf{W}_{i}$ represent the direction, density-related SH coefficients and latent feature-related SH coefficients at the $i$th point sampled on ray $\mathbf{r}$, respectively. \par

\paragraph{Full Objective Loss.} To learn the high-fidelity scene reconstruction, we optimize the following total loss in each iteration:
\begin{equation}
	\mathcal{L} = \mathcal{L}_{recon} + \lambda\mathcal{L}_{aniso}, 
\end{equation}
where $\lambda$ is a hyperparameter to scale the anisotropy regularization loss. In addition, to plug our anisotropy neural representation in different existing NeRFs for novel view synthesis, we remain the original losses and add the anisotropy regularization loss $\mathcal{L}_{aniso}$ in the full objective loss for model training.

\section{Experiments}
\label{sec:exper}

		In this section, we evaluate the effectiveness and generalizability of our method on novel view synthesis. We plug our anisotropic neural representation into existing state-of-the-art NeRFs and present quantitative and qualitative comparisons between the baseline NeRFs and our models on both synthetic and real-world benchmark datasets in Sec. \ref{comparison}. A comprehensive ablation study that supports our design choices is also provided in Sec. \ref{Ablation}. More details and per-scene results are provided in our supplementary materials.
		
		
		\paragraph{Datasets and metrics} 
		We report our results on two datasets: Blender \cite{mildenhall2021nerf} and Mip-360 \cite{barron2022mipnerf360}. Blender consists of eight synthetic scenes (\emph{lego, chair, hotdog, ficus, drums, materials, mic, and ship}), where each has 400 synthesized images. Mip-360 is an unbounded real-world dataset including three outdoor scenes (\emph{garden, bicycle, stump}) and four indoor scenes (\emph{room, kitchen, bonsai, counter}), and each scene contains a complex central object or region along with intricate background details. We follow the default split in \cite{barron2022mipnerf360} to produce training and testing views. In addition, we follow previous NeRF methods and report our quantitative results in terms of peak signal-to-noise ratio (PSNR), structural similarity index (SSIM) \cite{1284395}, learning perceptual image patch similarity (LPIPS) \cite{zhang2018unreasonable} and average error (Avg.) \cite{barron2021mipnerf} which summarizes three above metrics.

		
		
		\paragraph{Baselines.}
		We adopt the following four recently proposed neural rendering methods as baselines: two point-sampling methods Nerfacc \cite{li2023nerfacc} and K-Planes \cite{fridovich2023k}, and two shape-sampling methods Tri-Mip \cite{hu2023tri} and Zip-NeRF \cite{barron2023zip}. We use the official implementation of Nerfacc, K-Planes, and Tri-Mip and retrain the three models on the Blender and the PyTorch implementation of Zip-NeRF\cite{zipnerf_pytorch} and retrain Zip-NeRF and Nerfacc on Mip-360. 
		
		\paragraph{Implementation Details.} 
		For our anisotropic neural representation, we set the maximal degree of spherical harmonic basis $L=3$ and the hyperparameter $\lambda=1e^{-4}$. To keep the dimension of the predicted density $\mathbf{\sigma}$ and latent feature $\mathbf{e}$ the same as the baselines, we modify the output dimension of the first MLP $\mathcal{F_{\theta}}$ of the baselines to $(L+1)^2$ times the original output dimension. We keep other settings the same as our baselines. In addition, we use Nerfacc as our implementation backbone to verify our design choices in ablation studies. 
  \begin{table}[htb]
		\caption{Quantitative findings derived from the Blender dataset reveal significant advancements in average metrics with the incorporation of our method into the baselines.}
		
	\label{table:blender}
	\begin{center}
		{
				\setlength{\tabcolsep}{4mm}
				\begin{tabular}{lcccc}
					\toprule
					&  \textbf{PSNR}$\uparrow$ & \textbf{SSIM}$\uparrow$ &  \textbf{LPIPS}$\downarrow$  &\textbf{Avg.}$\downarrow$ \\
					\midrule
					Nerfacc &33.06&0.961&0.053&0.017\\
					Nerfacc+Ours &\textbf{34.08}&\textbf{0.966}&\textbf{0.046}&\textbf{0.015} \\
					\hline
					K-Planes &32.34&0.962& 0.052 & 0.018\\
					K-Planes+Ours &\textbf{33.00}&\textbf{0.964}&\textbf{0.052} & \textbf{0.017} \\
					\hline
					Tri-Mip &33.78&0.963&0.051&0.016\\
					Tri-Mip+Ours&\textbf{34.69}&\textbf{0.965}&\textbf{0.049}&\textbf{0.015}\\
					\bottomrule
				\end{tabular}
			}
		\end{center}
		
	\end{table}

		
		\begin{figure}[htb]
			\begin{center}
				\includegraphics[width=1\linewidth]{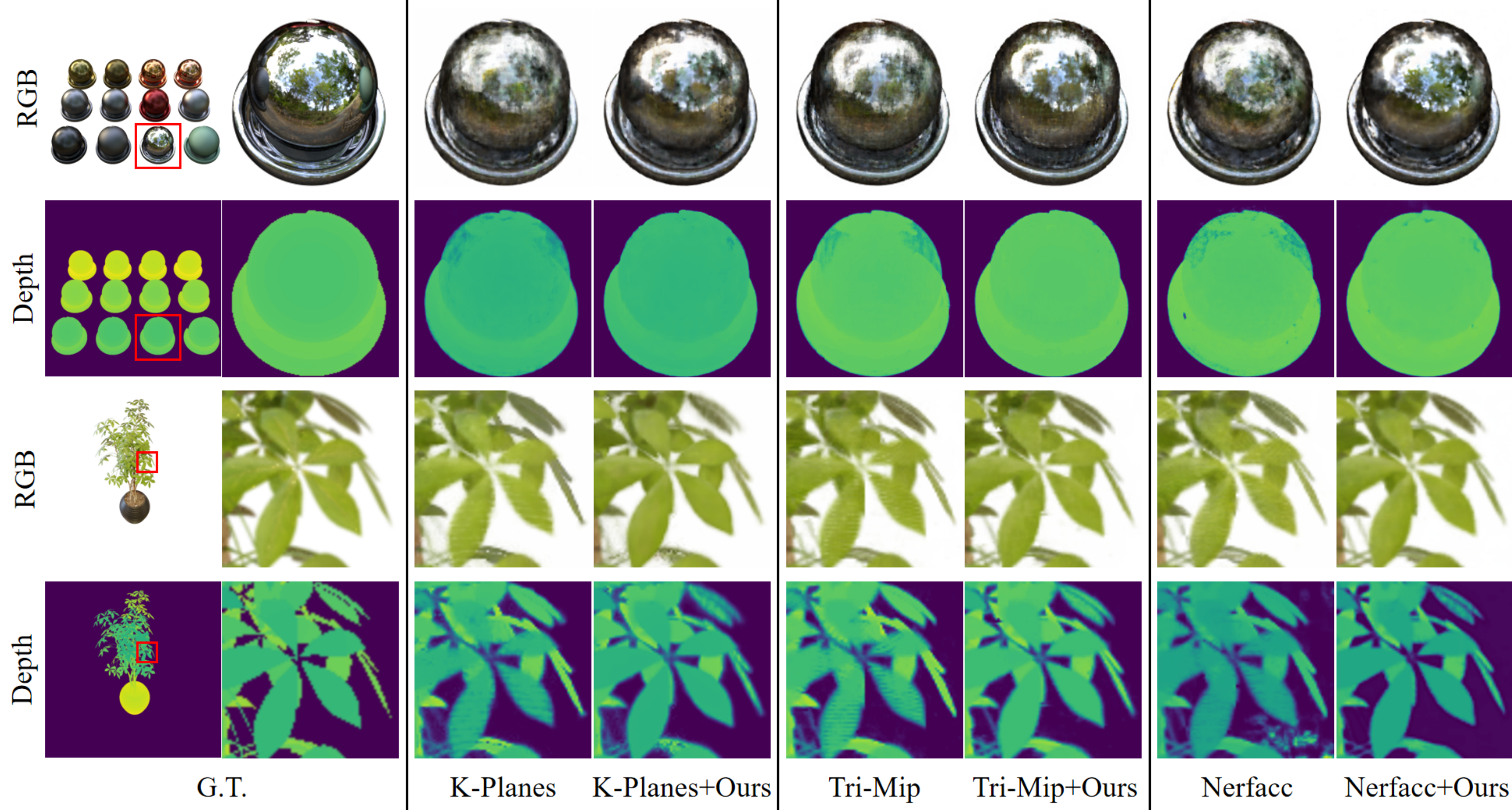}
			\end{center}
			\caption{Visual comparison of our method with baselines K-Planes\ \cite{fridovich2023k}, Tri-Mip\ \cite{hu2023tri} and Nerfacc\ \cite{li2023nerfacc} on synthetic scenes. Our representation works well on the three baselines, enabling to reduce blurring and artifacts and reconstruct better geometry and appearance on challenging details.}
			\label{fig:blender_result}
		\end{figure}

  \begin{table}[htb]
        \caption{
				Quantitative comparison results on Mip-360 dataset. Using our representation, the baseline models can achieve better rendering quality. }
			
			\label{table:M360_Result}
			\begin{center}
				{
					\setlength{\tabcolsep}{4mm}
					\begin{tabular}{lcccc}
						\toprule
						&\textbf{PSNR}$\uparrow$ & \textbf{SSIM}$\uparrow$ & \textbf{LPIPS}$\downarrow$ & \textbf{Avg.}$\downarrow$\\
						\midrule
						Nerfacc &27.47&0.771&0.294&0.063\\ 
						Nerfacc+Ours &\textbf{28.32}&\textbf{0.782}&\textbf{0.281}&\textbf{0.058}\\
						\hline
						Zip-NeRF&28.67& 0.838 & 0.270 & 0.053\\
						Zip-NeRF+Ours&\textbf{29.14}&\textbf{0.846}&\textbf{0.261} & \textbf{0.050}\\
						\bottomrule
				\end{tabular}}
			\end{center}
		\end{table}
  
		\begin{figure*}[htbp] 
                \begin{center}
				\includegraphics[width=1.0\linewidth]{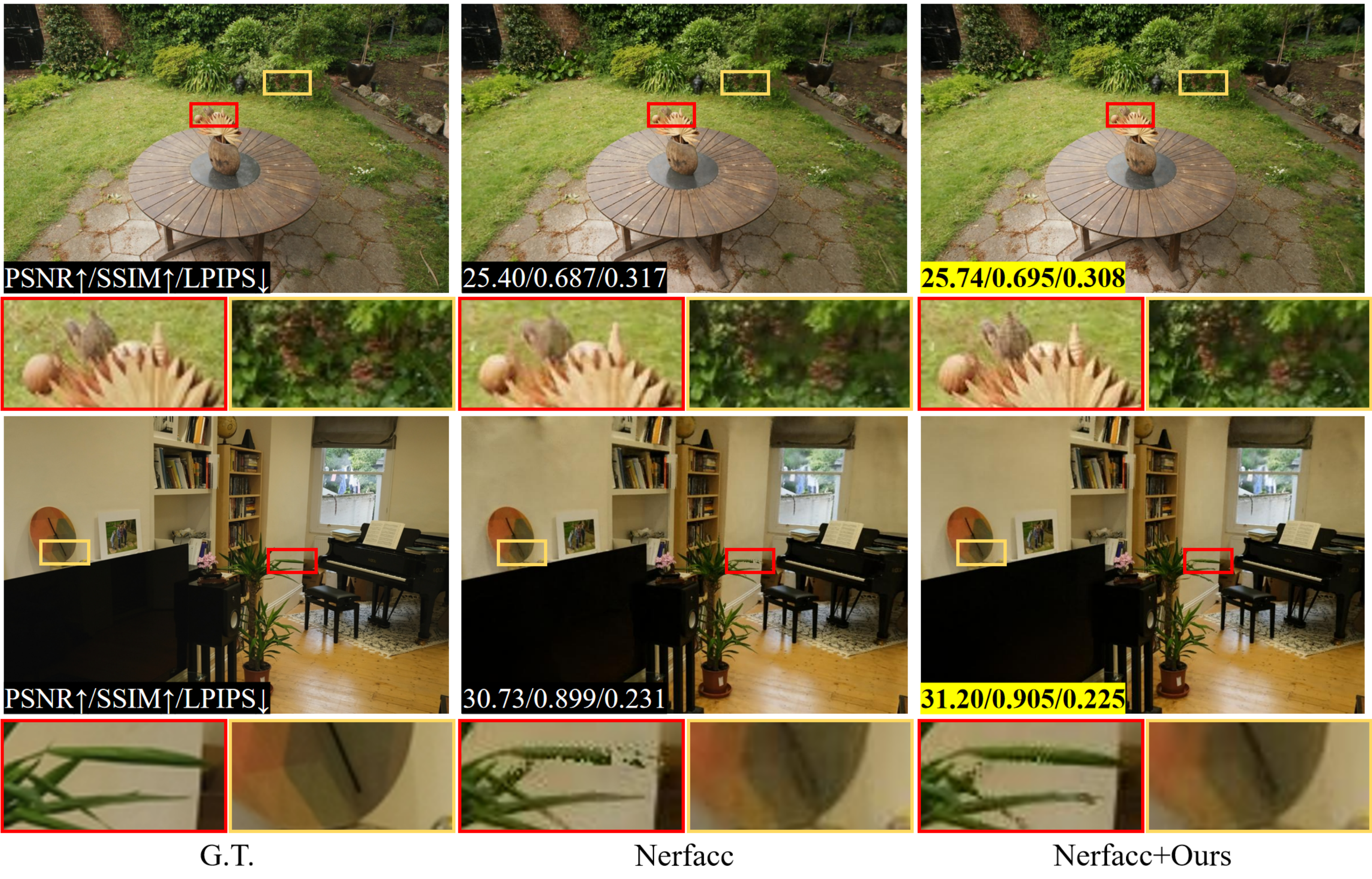}
			\end{center}
			\caption{
				Qualitative comparison of our method with Nerfacc on Mip-360. Our method enhances the rendering quality of vanilla Nerfacc, capturing more geometric details and producing more correct structure and texture.
			}
			\label{fig:m360_result}
                
                \vspace{\baselineskip} 
                
                \begin{center}
				\includegraphics[width=1.0\linewidth]{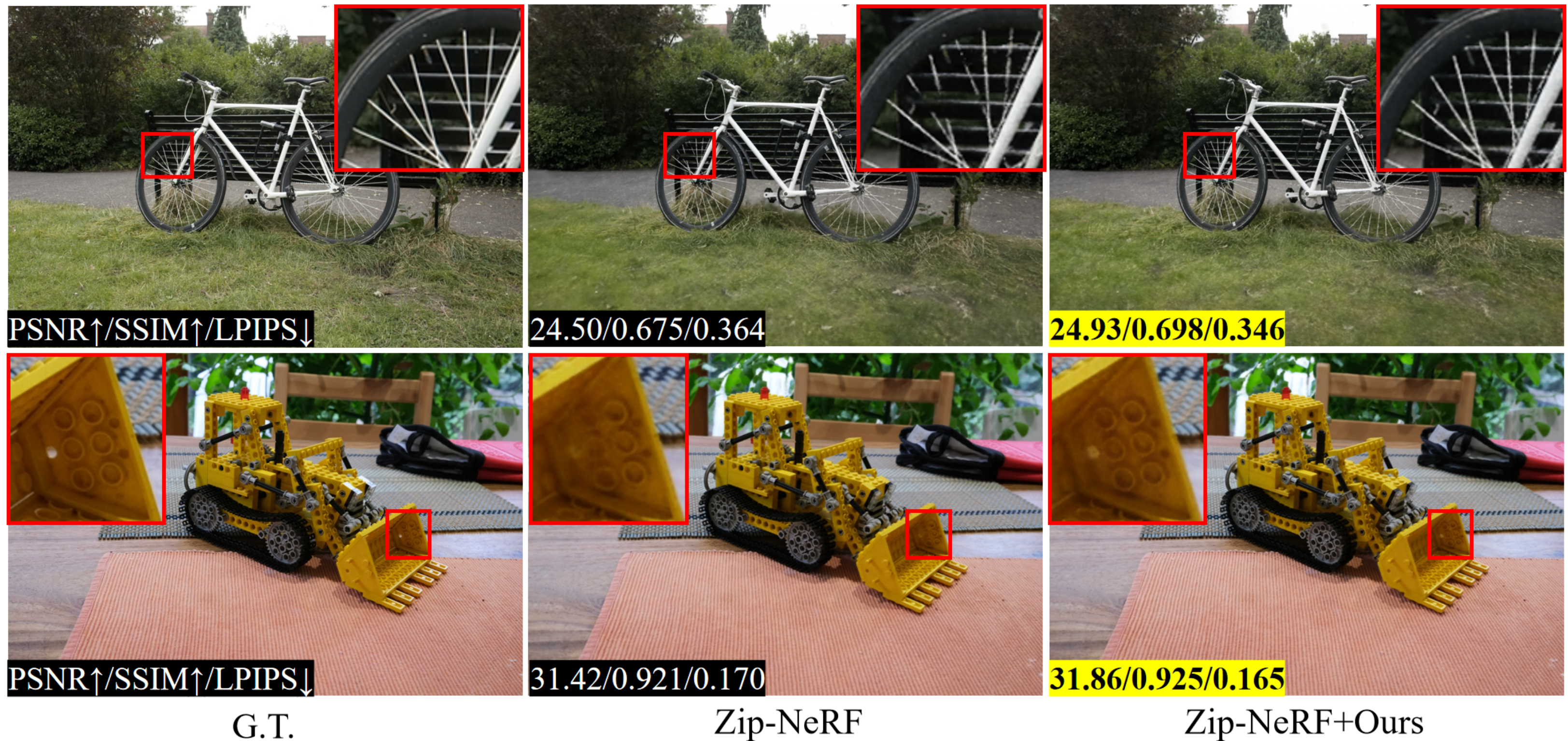}
			\end{center}
                \caption{Qualitative comparison of our method with Zip-NeRF on Mip-360. Our method enables them to estimate opacity more precisely and reconstruct finer details.}
			\label{fig:zip360result}
            \end{figure*}
		
		
  
\subsection{Comparisons}
\label{comparison}
      \paragraph{Quantitative Comparison.} Quantitative results on Blender are reported in Table \ref{table:blender}. We observe that our representation can significantly improve the numerical performance of the three baselines Nerfacc, K-Planes, and Tri-Mip. Quantitative results on Mip-360 are summarized in Table \ref{table:M360_Result}. 
      We see that our representation is also effective for real-world scenes. The rendering performance of Zip-NeRF+Ours indicates that our method can be used in combination with multi-sampling techniques to achieve further better rendering details.
		\paragraph{Qualitative Comparison.} Fig. \ref{fig:blender_result} shows the visual comparison and depth results of two synthetic scenes, including \emph{materials} and \emph{ficus} scenes in Blender. It shows that our method can assist the baseline models in recovering the finer appearance and geometric details with less blur and artifacts, such as reflections and concentrated surface of \emph{materials}, tiny leaves, and semi-translucent edges of \emph{ficus}. The rendering results of real-world scenes are illustrated in Fig. \ref{fig:m360_result} and  Fig. \ref{fig:zip360result}. As shown in Fig. \ref{fig:m360_result}, although the vanilla Nerfacc can reconstruct the overall scene geometry well, the tiny objects and local textures are blurry. Our method can help the model capture more geometric details like the leaves and the potted plants in the \emph{garden}, and generate better appearance, such as cleaner leaves and plates in the \emph{room}. Fig. \ref{fig:zip360result} shows that Zip-NeRF renders photo-realistic images but fails to reconstruct some challenging small structures like banners of \emph{bicycle} and lego of \emph{kitchen}. Our method can improve the visual performance of Zip-NeRF, producing sharper and more accurate renderings.

            \begin{table}[htb]
                \caption{Quantitative ablation study of the design choices of our anisotropic neural representation on Mip-360 and Blender.}
			\label{table:ablation_anr}
			\begin{center}
				{
					{
						\begin{tabular}{lcccccccc}
							\toprule
							& \multicolumn{4}{c}{\textbf{Blender}} &\multicolumn{4}{c}{\textbf{Mip-360}}\\
							\cmidrule(lr){2-5}\cmidrule(lr){6-9}
							 & \textbf{PSNR}$\uparrow$ &\textbf{SSIM}$\uparrow$ & \textbf{LPIPS}$\downarrow$& \textbf{Avg.}$\downarrow$& \textbf{PSNR}$\uparrow$&\textbf{SSIM}$\uparrow$& \textbf{LPIPS}$\downarrow$& \textbf{Avg.}$\downarrow$ \\
							\midrule
							Nerfacc &33.06&0.961&0.053&0.017&27.47&0.771&0.294&0.063\\
							Nerfacc+aniso-$\sigma$ &33.41&0.961&0.050&0.017&28.07&0.774&0.290&0.060\\
							Nerfacc+aniso-$\mathbf{e}$&33.68&0.963&0.049&0.016&28.21&0.779&0.286&0.059\\
							Nerfacc+Ours &\textbf{34.08}&\textbf{0.966}&\textbf{0.046}&\textbf{0.015}&\textbf{28.32}&\textbf{0.782}&\textbf{0.281}&\textbf{0.058}\\
							\bottomrule
						\end{tabular}
					}
				}
			\end{center}
		\end{table}


	\subsection{Ablation Study}
	\label{Ablation}
	\paragraph{Anisotropic neural representation.}
		We first verify the effectiveness of our anisotropic neural representation. In Table \ref{table:ablation_anr}, we compare our proposed representation to two variations and a baseline (\emph{Nerfacc}). The first one (\emph{Nerfacc+aniso-$\sigma$}) is to replace the anisotropic latent feature $\mathbf{e}(\mathbf{d}, \mathbf{W})$ with isotropic one $\mathbf{e}$ and uses only anisotropic density $\mathbf{\sigma}(\mathbf{d}, \mathbf{k})$. This change with anisotropic density can improve the rendering quality of the vanilla Nerfacc but it leads to large performance drops on all metrics compared to our design, showing the necessity of having the anisotropic latent feature for rendering. The second one (\emph{Nerfacc+aniso-$\mathbf{e}$}) is to replace the anisotropic density $\mathbf{\sigma}(\mathbf{d}, \mathbf{k})$ with the isotropic density $\mathbf{\sigma}$ and uses only the anisotropic latent feature $\mathbf{e}(\mathbf{d}, \mathbf{W})$. Although this also enhances the rendering quality of the vanilla Nerfacc, it produces inferior performance than our overall design, indicating that the anisotropic density can indeed help to learn geometry better. Additionally, The second variation (\emph{Nerfacc+aniso-$\mathbf{e}$}) yields slightly better numerical performance than the first variation (\emph{Nerfacc+aniso-$\sigma$}). This can be explained by the fact that the dimension of the anisotropic latent feature $\mathbf{e}(\mathbf{d}, \mathbf{W})$ is higher than the density $\mathbf{\sigma}(\mathbf{d}, \mathbf{k})$, which brings more performance gains. 
        \begin{figure}[tb]
        	\begin{center}
        	\includegraphics[width=1\linewidth]{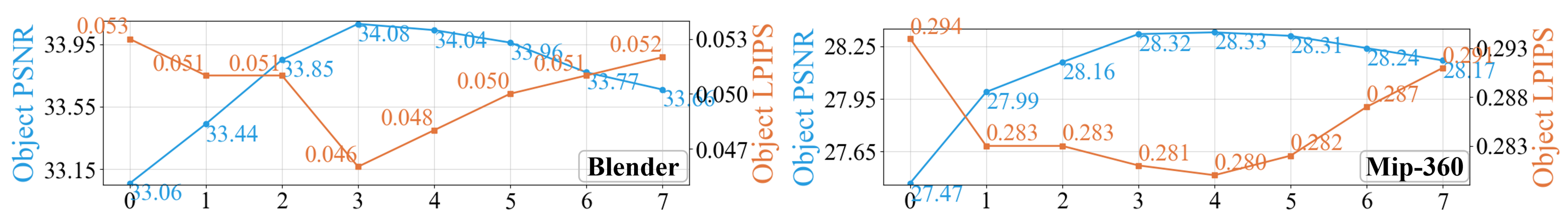}
        		\end{center}
        		\caption{The impact of the maximal degree of SH functions on rendering quality in terms of PSNR and LPIPS. The $x$-axis refers to the maximal SH degree $L$.}
        		\label{fig:ablation1}
        \end{figure}
	\paragraph{Maximal Degree of SH basis.}
		In Fig. \ref{fig:ablation1}, we show the impact of the maximal degree of SH basis on rendering quality in terms of PSNR and LPIPS. Our model is the same as vanilla Nerfacc when setting $L=0$. Our model gets the best rendering quality when setting $L=3$ on Blender and $L=4$ on Mip-360, since higher maximal degree more anisotropy can be captured by SH functions but continuously increasing the maximal degree will lead to anisotropy over-fitting and can hardly be interpolated accurately. Meanwhile, the computation complexity of SH functions ($O(L^2)$) increases exponentially as $L$ increases. We set $L=3$ in our experiments.


                \begin{table}[htb]
                \caption{Quantitative ablation study of anisotropy regularization loss on Mip-360 and Blender.}
			\label{table:ablation_re}
			\begin{center}
				{
					{
						\begin{tabular}{lcccccccc}
							\toprule
							& \multicolumn{4}{c}{\textbf{Blender}} &\multicolumn{4}{c}{\textbf{Mip-360}}\\
							\cmidrule(lr){2-5}\cmidrule(lr){6-9}
							 & \textbf{PSNR}$\uparrow$ &\textbf{SSIM}$\uparrow$ & \textbf{LPIPS}$\downarrow$& \textbf{Avg.}$\downarrow$& \textbf{PSNR}$\uparrow$&\textbf{SSIM}$\uparrow$& \textbf{LPIPS}$\downarrow$& \textbf{Avg.}$\downarrow$ \\
							\midrule
                                w/o $\mathcal{L}_{aniso}$.&33.69&0.962&0.051&0.016&28.01&0.775&0.285&0.060\\
                                w/o aniso-$\sigma$ in $\mathcal{L}_{aniso}$ &33.88&0.964&0.049&0.016&28.13&0.775&0.285&0.059\\
							w/o aniso-$\mathbf{e}$ in $\mathcal{L}_{aniso}$ &33.81&0.963&0.050&0.016&28.25&0.778&0.284&0.059\\
							Nerfacc+Ours &\textbf{34.08}&\textbf{0.966}&\textbf{0.046}&\textbf{0.015}&\textbf{28.32}&\textbf{0.782}&\textbf{0.281}&\textbf{0.058}\\
							\bottomrule
						\end{tabular}
					}
				}
			\end{center}
		\end{table}

        \begin{figure}[htb]
        	\begin{center}
        	\includegraphics[width=0.95\linewidth]{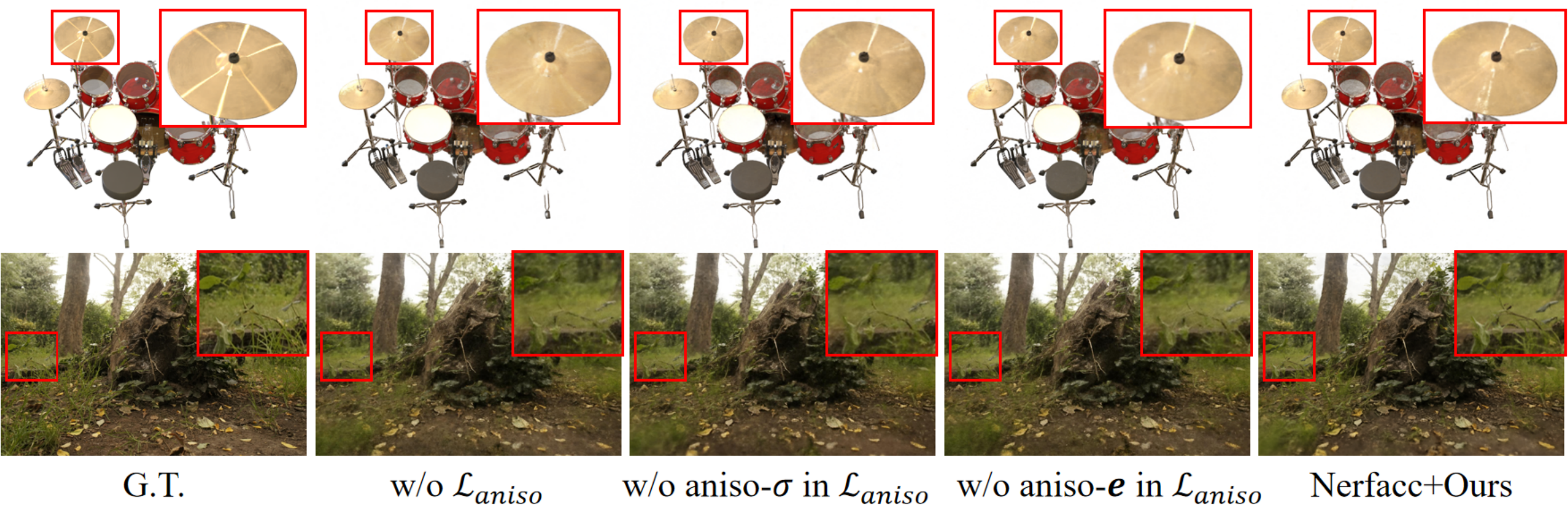}
        		\end{center}
        		\caption{Qualitative ablation study of our anisotropy regularization loss. Using our full loss with $\mathcal{L}_{recon}$ and $\mathcal{L}_{aniso}$ renders novel views with sharper texture details and more accurate geometry.}
        		\label{fig:ablation_pic}
          \end{figure}

     \begin{figure}[htb]
        	\begin{center}
        	\includegraphics[width=1\linewidth]{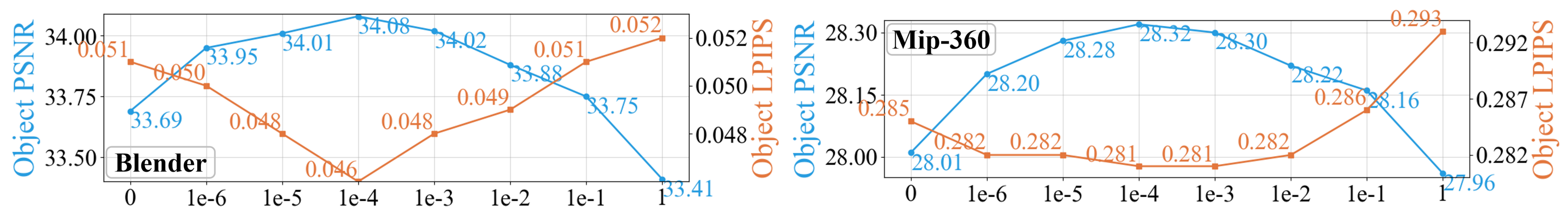}
        		\end{center}
        		\caption{The effect of the strength of anisotropy regularization $\lambda$ on rendering quality. The $x$-axis refers to $\lambda$.}
        		\label{fig:ablation2}
        \end{figure}

        \paragraph{Anisotropy regularization.}
        We first verify the effectiveness of different anisotropy regularization loss terms for the mapping process from Sec. \ref{subsec: training}. The experimental results in Table \ref{table:ablation_re} and Fig. \ref{fig:ablation_pic} show that using all losses together leads to the best overall performance. Without anisotropy regularization loss (\emph{w/o $\mathcal{L}_{aniso}$}) or without anisotropy density (\emph{w/o aniso-$\sigma$}) or latent feature (\emph{w/o aniso-$\mathbf{e}$}) regularization in $\mathcal{L}_{aniso}$, the reconstruction performance drops significantly, indicating that these anisotropy regularization losses are important for the disambiguation of the optimization process. We also investigate the impact of anisotropy regularization strength in Fig. \ref{fig:ablation2}. Our method benefits more from a larger $\lambda$ in terms of PSNR and LPIPS across two datasets, with $\lambda=1e^{-4}$ being the best. Note that $\mathcal{L}_{aniso}$ is disabled when $\lambda=0$. When $\lambda \geq1e^{-3}$, the rendering quality decreases. Because the overly strong anisotropy penalization leads to limited anisotropy in representation. In our experiments, we set $\lambda=1e^{-4}$ to balance anisotropy capturing and over-fitting. 
 
  

			

\section{Conclusion}
\label{sec: conclu}
We introduced a novel anisotropic neural representation for NeRFs using spherical harmonic (SH) functions, which enables accurate representation and novel view rendering in complex scenes. We used SH functions and the corresponding coefficients that were estimated by a modified NeRF MLP to determine the anisotropic features. During training, an anisotropy regularization is introduced to alleviate the anisotropy over-fitting problem. We showed that our design results in a simple and flexible representation module that is easy to generalize to various NeRFs. Experiments on both synthetic and real-world datasets demonstrated the effectiveness of our method in improving the rendering quality for NeRFs. 

%
%
\bibliographystyle{splncs04}
\bibliography{main}
\end{document}